\documentclass[final]{cvpr}

\usepackage{times}
\usepackage{epsfig}
\usepackage{graphicx}
\usepackage{amsmath}
\usepackage{amssymb}
\usepackage{multirow}
\usepackage{color}
\usepackage{booktabs}
\usepackage{bbding}
\usepackage{makecell}

\usepackage[pagebackref=true,breaklinks=true,colorlinks,bookmarks=false]{hyperref}

\begin{document}

\setlength{\parskip}{0pt}

\title{PP-PicoDet: A Better Real-Time Object Detector on Mobile Devices}  

\author{Guanghua Yu, Qinyao Chang, Wenyu Lv, Chang Xu, Cheng Cui, Wei Ji, \\
Qingqing Dang, Kaipeng Deng, Guanzhong Wang, Yuning Du, \\
Baohua Lai, Qiwen Liu, Xiaoguang Hu, Dianhai Yu, Yanjun Ma \\
Baidu Inc.\\
\tt\small \{yuguanghua02, changqinyao, lvwenyu01, dangqingqing, dengkaipeng\} @baidu.com
}

\maketitle
\thispagestyle{empty}

\begin{abstract}

The better accuracy and efficiency trade-off has been a challenging problem in object detection. In this work, we are dedicated to studying key optimizations and neural network architecture choices for object detection to improve accuracy and efficiency. We investigate the applicability of the anchor-free strategy on lightweight object detection models. We enhance the backbone structure and design the lightweight structure of the neck, which improves the feature extraction ability of the network. We 
improve label assignment strategy and loss function to make training more stable and efficient. Through these optimizations, we create a new family of real-time object detectors, named PP-PicoDet, which achieves superior performance on object detection for mobile devices. Our models achieve better trade-offs between accuracy and latency compared to other popular models. PicoDet-S with only 0.99M parameters achieves 30.6\% mAP, which is an absolute 4.8\% improvement in mAP while reducing mobile CPU inference latency by 55\% compared to YOLOX-Nano, and is an absolute 7.1\% improvement in mAP compared to NanoDet. It reaches 123 FPS (150 FPS using Paddle Lite) on mobile ARM CPU when the input size is 320. PicoDet-L with only 3.3M parameters achieves 40.9\% mAP, which is an absolute 3.7\% improvement in mAP and 44\% faster than YOLOv5s. As shown in Figure \ref{figure1}, our models far outperform the state-of-the-art results for lightweight object detection. Code and pre-trained models are available at PaddleDetection$\footnote{\scriptsize\url{https://github.com/PaddlePaddle/PaddleDetection}\label{
ppdet1}}$.

\end{abstract}

\section{Introduction}
Object detection is widely adopted in numerous computer vision tasks, including autonomous driving, robot vision, intelligent transportation, industrial quality inspection, object tracking, etc. 
Two-stage models normally lead to higher performance. However, this type of resource-consuming network limits the adoption of real-world applications. To overcome this problem, lightweight mobile object detectors have attracted increasing research interests aiming to design highly efficient object detection.
Modern object detectors in the YOLO series have \cite{redmon2018yolov3,bochkovskiy2020yolov4,glenn_jocher_2021_5563715,ge2021yolox} become popular since they are in a small subset of works that consider resource constraints. Compared to two-stage models, the YOLO series has better efficiency and high accuracy as well. However, the YOLO series does not deal with the following problems: 1) the need to carefully and manually re-design anchor boxes to adopt different datasets. 2) the problem of imbalance between positive and negative samples as most of the generated anchors are negative.

\begin{figure}[t]
\centering
\includegraphics[width=\linewidth]{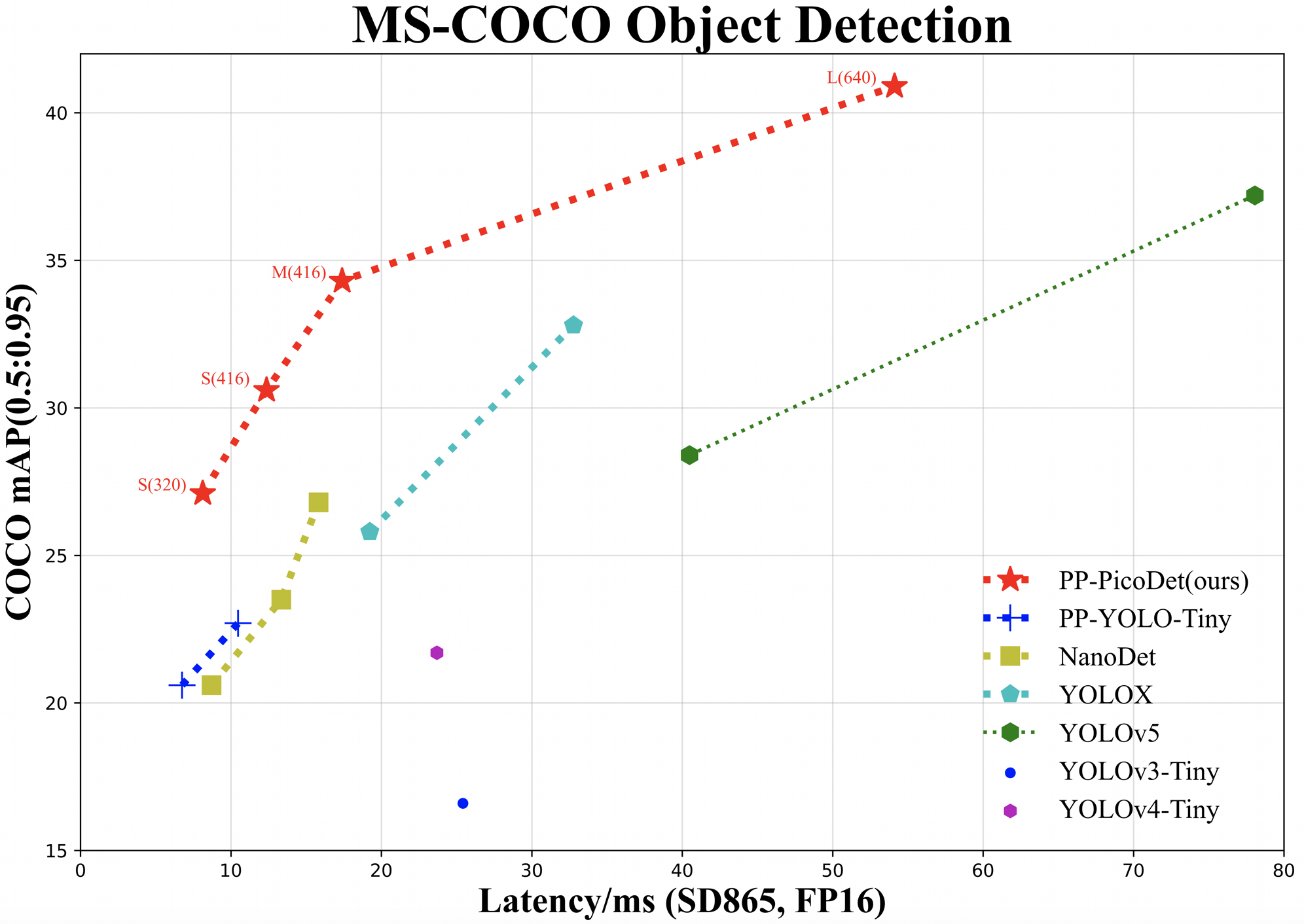} 
\caption{Comparison of the mAPs of different lightweight models. The latency of all models tested on Qualcomm Snapdragon$^\circledR$ 865(4*A77+4*A55) Processor with batch size of 1. The details are presented in Table \ref{tab-state-of-the-art}.}
\label{figure1}
\end{figure}

In recent years, many works have aimed to develop more efficient detector architectures, such as anchor-free detectors. FCOS\cite{tian2019fcos} solves the problem of overlaps within the ground-truth labels. There is no complicated hyper-parameter tuning compared to other anchor-free detectors. However, most anchor-free detectors are large-scale server detectors. In the small minority, NanoDet\cite{rangiLyu2021nanodet} and YOLOX-Nano\cite{ge2021yolox} are the anchor-free detectors and also mobile detectors. The problem is that lightweight anchor-free detectors usually cannot balance the accuracy and efficiency well. So in this work, inspired by FCOS and GFL\cite{li2020generalized}, we propose an improved mobile-friendly and high-accuracy anchor-free detector named PP-PicoDet. To summarize, our main contributions are as follows:

\begin{itemize}
\item[$\bullet$]
We adopt the CSP structure to construct CSP-PAN as the neck. The CSP-PAN unifies the input channel numbers by $1\times1$ convolution for all branches of the neck, which significantly enhances the feature extraction ability and reduces network parameters. And we enlarge $3\times3$ depthwise separable convolution to $5\times5$ depthwise separable convolution to expand the receptive field.
\end{itemize}
\begin{itemize}
\item[$\bullet$] The label assignment strategy is essential in object detection. We use SimOTA\cite{ge2021yolox} dynamic label assignment strategy and optimize some calculation details. Specifically, we use the weighted sum of Varifocal Loss (VFL)\cite{zhang2021varifocalnet} and GIoU loss\cite{zheng2020distance} to calculate the cost matrix, enhancing accuracy without harming efficiency.
\end{itemize}
\begin{itemize}
\item[$\bullet$] ShuffleNetV2\cite{shufflenetv2} is cost-effective on mobile devices. We further enhance the network structure and propose a new backbone, namely Enhanced ShuffleNet (ESNet), which performs better than ShuffleNetV2.
\end{itemize}
\begin{itemize}
\item[$\bullet$] We propose an improved detection One-Shot Neural Architecture Search (NAS) pipeline to find the optimal architecture automatically for object detection. We straightly train the supernet on detection datasets, which leads to significant computational savings and optimization for detection. Our NAS-generated models achieve better efficiency and accuracy trade-offs.
\end{itemize}

Through the above optimizations, we propose a series of models that far outperform the state-of-the-art results of lightweight object detection. As shown  in Table \ref{tab-state-of-the-art}, PicoDet-S achieves 30.6\% mAP with only 0.99M parameters and 1.08G FLOPs. It achieves 150 FPS on mobile ARM CPU when the input size is 320. PicoDet-M achieves 34.3\% mAP with only 2.15M parameters and 2.5G FLOPs. PicoDet-L achieves 40.9\% mAP with only 3.3M parameters and 8.74G FLOPs. We provide small, medium, and large models to support different deployments. All our experiments are implemented based on PaddlePaddle$\footnote{\url{https://github.com/PaddlePaddle}}$. Code and pre-trained models are available at PaddleDetection\cite{ppdet2021}.

\begin{figure*}[t]
\centering
\includegraphics[width=1.0\textwidth]{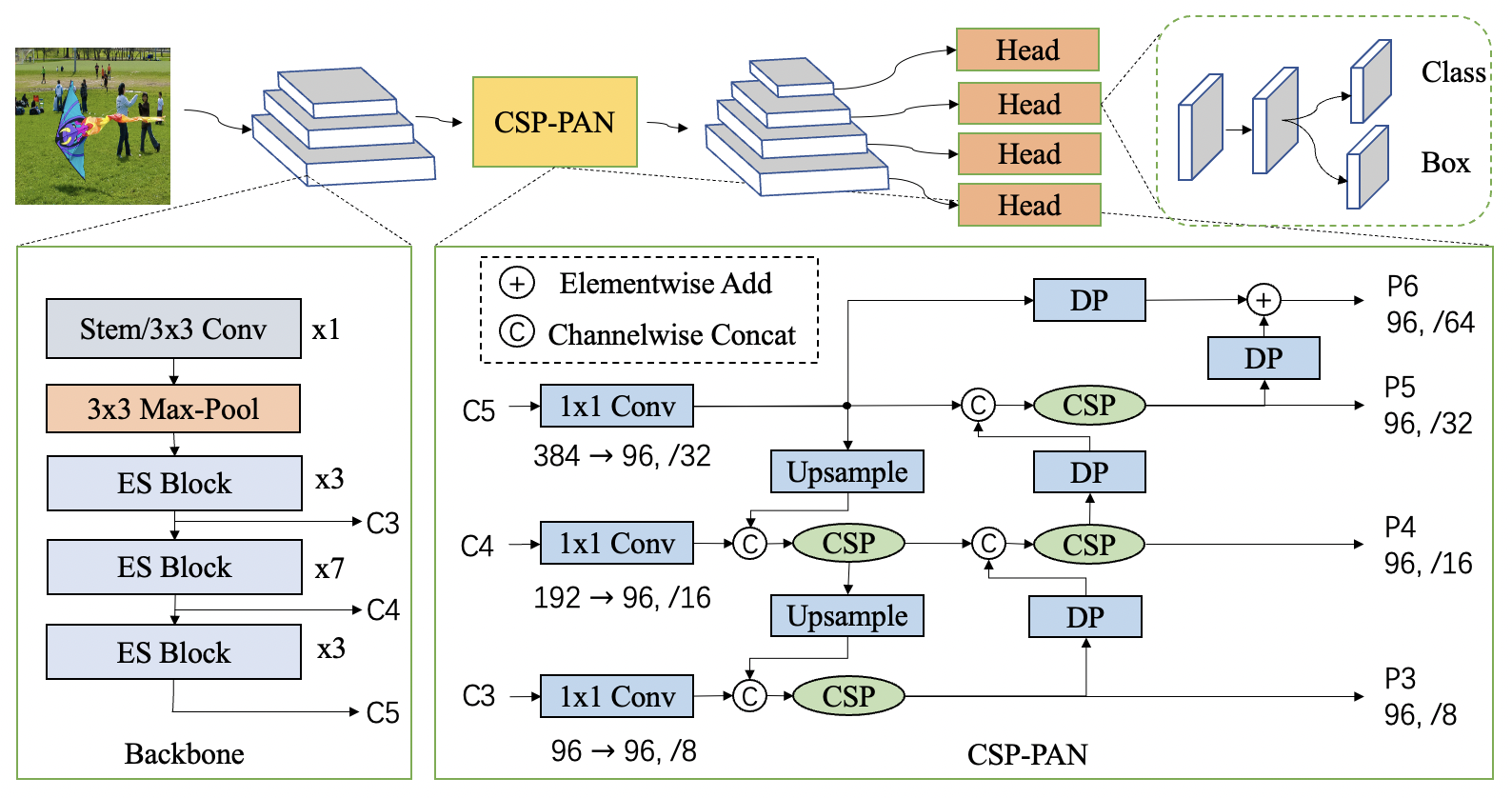} 
\caption{PP-PicoDet Architecture. The Backbone is ESNet, which outputs C3-C5 feature maps to the neck. The neck is CSP-PAN, which inputs three feature maps and outputs four feature maps. For PP-PicoDet-S, the input channel numbers are [96, 192, 384], and  the output channel numbers are [96, 96, 96, 96]. DP module uses depthwise and pointwise convolution.}
\label{Picodet}
\end{figure*}


\section{Related Works}

Object detection is a classic computer vision challenge that aims to identify the object category and object location in pictures or videos. Existing object detectors, can be divided into two categories: anchor-based detectors, and anchor-free detectors., The two-stage detectors\cite{girshick2015fast,he2015spatial,fasterrcnn,fpn,maskrcnn,cai2018cascade} which are normally anchor-based, generate region proposals from the image and then generate the final bounding box from the region proposals. To improve the accuracy of object positioning, FPN\cite{fpn} fuses multi-scale high-level semantic features. The two-stage detectors are more accurate in the object positioning, while it is difficult to achieve real-time detection on the CPU or ARM devices. The one-stage object detectors\cite{liu2016ssd,redmon2017yolo9000,redmon2018yolov3,bochkovskiy2020yolov4,glenn_jocher_2021_5563715,long2020pp,huang2021pp,ge2021yolox} are also anchor-based detectors, which have a better balance between speed and accuracy, thus have been widely used in practice. SSD\cite{liu2016ssd}, detecting multiple-scale objects, is more friendly to small objects, but it's not competitive in accuracy. At the same time, the YOLO series(except YOLOv1\cite{redmon2016you}) performs well in both accuracy and speed. However, it does not tackle some problems we analyzed in the previous section.

The anchor-free detectors\cite{law2018cornernet,duan2019centernet,tian2019fcos} aim to eliminate the anchor boxes, which is a significant improvement in object detection. The main idea of YOLOv1 is to divide the image into multiple grids and then predict bounding boxes at points near the center of objects. CornerNet\cite{law2018cornernet} detects a pair of corners of a bounding box without designing the anchor boxes as priori boxes. CenterNet\cite{zhou2019objects} abandons the upper left corner and the bottom right corner, directly detecting the center point. FCOS\cite{tian2019fcos} first reformulates object detection in a per-pixel prediction fashion and proposes a “centerness” branch. The anchor-free detectors solve some problems of the anchor-based detectors, which reduce the memory cost and provide a more accurate calculation of the bounding box.

Later works further improve the object detector from different aspects. ATSS\cite{atss} proposes an adaptive training sample selection to automatically select positive and negative samples according to the statistical characteristics of objects. Generalized Focal Loss(GFL)\cite{li2020generalized} eliminates the "centerness" branch in FCOS and merges the quality estimation into the class prediction vector to form a joint representation of localization quality and classification. 

In the field of mobile object detection, a lot of effort has been devoted to achieving more accurate and efficient object detectors. Through compression-compilation collaborative design of YOLOv4\cite{bochkovskiy2020yolov4}, YOLObile\cite{cai2020yolobile} realizes real-time object detection on mobile devices. PP-YOLO-Tiny\cite{ppdet2021} adopts MobileNetV3\cite{mbv3} backbone and TinyFPN structure based on PP-YOLO\cite{long2020pp}. NanoDet\cite{rangiLyu2021nanodet} uses ShuffleNetV2\cite{shufflenetv2} as its backbone to make the model lighter and uses ATSS and GFL to enhance accuracy. YOLOX-Nano is currently the lightest model in the YOLOX\cite{ge2021yolox} series, using dynamic label assignment strategy SimOTA to achieve their best performance within acceptable parameters.

Hand-crafted technologies heavily rely on expert knowledge and tedious trials. In recent years, NAS has shown promising results discovering and optimizing network architectures, e.g., MobileNetV3, EfficientNet\cite{tan2019efficientnet}, and Mnasnet\cite{tan2019mnasnet}. NAS thus can be an excellent choice to generate a detector with a better efficiency-accuracy trade-off. One-shot NAS methods save computational resources by sharing the same weights mutually. Numerous One-shot NAS works on image classification in recent years, e.g., ENAS\cite{pham2018efficient}, SMASH\cite{brock2017smash}. To our best knowledge, fewer attempts have been made to develop NAS for object detection. NAS-FPN\cite{ghiasi2019fpn} searches for feature pyramid networks (FPN). DetNas\cite{chen2019detnas} firstly trains the supernet backbone on ImageNet and then finetunes the supernet on COCO. MobileDets\cite{xiong2021mobiledets} use NAS and propose an augmented search space family to achieve better latency-accuracy trade-off on mobile devices.

\section{Approach}

In this section, we first present our design ideas and NAS search method of better backbone, which help us to improve accuracy and reduce latency. Then, we offer enhanced strategies of the neck and head modules. Finally, we describe the label assignment strategy and other strategies to improve the performance further.

\subsection{Better Backbone}
\begin{figure}[t]
\centering
\includegraphics[width=\linewidth]{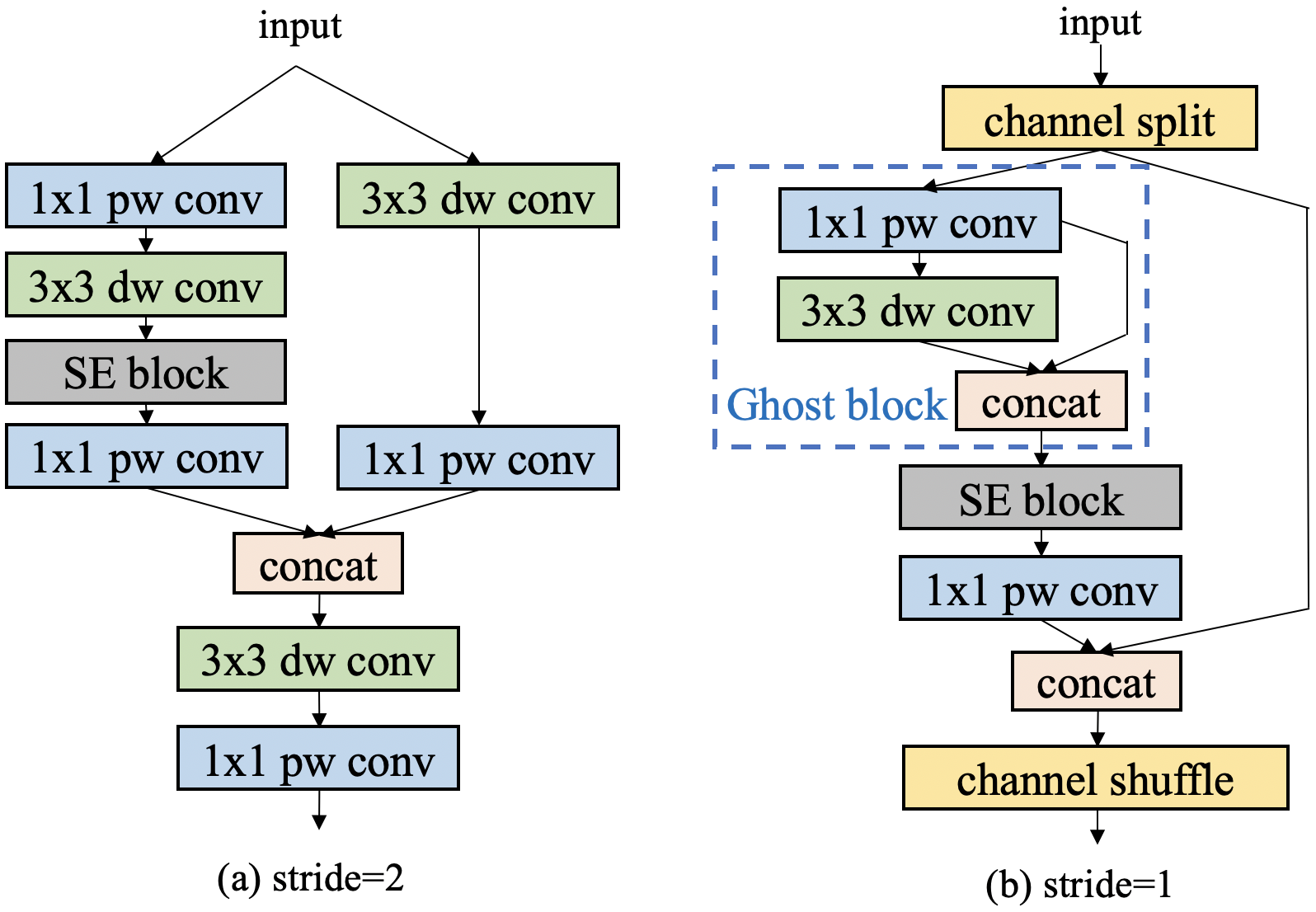} 
\caption{ES Block architecture. (a) ES Block with stride=2; (b) ES Block with stride=1.}
\label{figure-es-block}
\end{figure}

\textbf{Manually Designed Backbone.} 
Based on many experiments, we find that ShuffleNetV2 is more robust than other networks on mobile devices. To further improve the performance of ShuffleNetV2, we follow some methods of PP-LCNet\cite{cui2021pplcnet} to enhance the network structure and build a new backbone, namely Enhanced ShuffleNet (ESNet). Figure \ref{figure-es-block} describes the ES Block of ESNet in detail. 
The SE module\cite{senet} does a good job of weighting the network channels for better features. Therefore, we add SE modules to all blocks. Like MobileNetV3, the activation functions for the two layers of the SE module are ReLU and H-Sigmoid, respectively. Channel shuffle provides the information exchange of ShuffleNetV2 channels, but it causes the loss of fusion features. To address this problem, we add depthwise convolution and pointwise convolution to integrate different channel information when the stride is 2 (Figure \ref{figure-es-block}a). The author of GhostNet\cite{ghostnet} proposes a novel Ghost module that can generate more feature maps with fewer parameters to improve the network's learning ability. We add the Ghost module in the blocks with stride set to 1 to further enhance the performance of our ESNet (Figure \ref{figure-es-block}b). 

\textbf{Neural Architecture Search.} 
At the same time, we present the first effort on one-shot searching for object detectors. Object detectors, equipped with high-performance backbones for classification, might be sub-optimal due to the gap between different tasks. We do not search for a better classifier, but train and search the detection supernet directly on the detection datasets, which leads to significant computational savings and optimization of detection instead of classification. The framework simply consists of two steps: (1) training the one-shot supernet on detection datasets, (2) architecture search on the trained supernet with an evolutionary algorithm (EA).

For convenience, we simply use channel-wise search for backbone here. Specifically, we give flexible ratio options to choose different channel ratios. We choose the ratio randomly and coarsely in [0.5, 0.675, 0.75, 0.875, 1]. For example, 0.5 represents that the width is scaled by 0.5 of the full model. The channel numbers divisible by 8 can improve the speed of inference time on hardware devices. Therefore, instead of using the channel numbers in the original model, we first trained the full model with channel numbers [128, 256, 512] for each stage block. All ratio options can also keep the channel numbers divisible by 8. The chosen ratio works on all prunable convolutions in each block. All output channels are fixed as the full model. To avoid tedious hyper-parameter tuning, we 
fix all original settings in the architecture search. For the training strategy, we adopt the sandwich rule to sample the largest (full) and the smallest child model and six randomly sampled child models for each training iteration. There are no more additional techniques adopted in the training strategy, such as distillation, since different techniques perform inconsistently for different models, especially for detection tasks. Finally, the selected architectures are retrained on ImageNet dataset\cite{imagenet} and then 
trained on COCO\cite{mscoco}. 

\subsection{CSP-PAN and Detector Head}
We use the PAN\cite{liu2018path} structure to obtain multi-level feature maps and the CSP structure for feature concatenation and fusion between the adjacent feature maps. The CSP structure is widely used in the neck of YOLOv4\cite{bochkovskiy2020yolov4} and YOLOX\cite{ge2021yolox}. In the original CSP-PAN, the channel number in each output feature map is kept the same as the input from the backbone. The structures with large channel numbers have expensive computational costs for mobile devices. We address this problem through making all channel numbers in all feature maps equal to the smallest channel number by $1\times1$ convolution. Top-down and bottom-up feature fusion are then used through the CSP structure. The scaled-down features lead to lower computation costs and undamaged accuracy. Furthermore, we add a feature map scale to the top of CSP-PAN to detect more objects. At the same time, all convolutions except $1\times1$ convolutions are depthwise separable convolution. Depthwise separable convolution expands the receptive field through $5\times5$ convolution. This structure brings a considerable increase in accuracy with much fewer parameters. The specific structure is shown in the Figure \ref{Picodet}.

In the detector head, we use depthwise separable convolution and $5\times5$ convolution to expand the receptive field. The numbers of depthwise separable convolution can be set to 2, 4, or more. The overall network structure is shown in the Figure \ref{Picodet}. Both the neck and the head have four scale branches. We keep the channel numbers in the head consistent with the neck module and couple the classification and regression branches. YOLOX\cite{ge2021yolox} uses a decoupled head with fewer channel numbers to improve accuracy. Our coupled head performs better without reducing the channel numbers. The parameters and the inference speed are almost the same as the decoupled head.

\begin{table*}[!htbp]
\centering
\begin{center}
\begin{tabular}{c|c|c|c|c|c|c}
\toprule[1pt]
Model & Size & Params(M) & FLOPs(G) & mAP(0.5:0.95) & mAP(0.5) & Latency(ms) \\
\midrule[1pt]
YOLOv3-Tiny & 416   & 8.86  & 5.62  & 16.6  & 33.1   & 25.42 \\
YOLOv4-Tiny & 416   & 6.06  & 6.96  & 21.7  & 40.2   & 23.69 \\
MobileDet-CPU & 320   & 3.85  & 1.02  & 24.2  & -   & - \\
YOLObile & 320   & 4.59  & 3.59  & 31.6  & 49.0   & - \\
\hline
PP-YOLO-Tiny & 320   & 1.08  & 0.58  & 20.6  & -   & 6.75 \\
PP-YOLO-Tiny & 416   & 1.08  & 1.02  & 22.7  & -   & 10.48 \\
\hline
NanoDet-M   & 320   & 0.95     & 0.72     & 20.6	 & -   & 8.71 \\
NanoDet-M   & 416   & 0.95     & 1.2     & 23.5	 & -   & 13.35 \\
NanoDet-M-1.5x  & 416   & 2.08     & 2.42     & 26.8	 & -   & 15.83 \\
\hline
YOLOX-Nano  & 416   & 0.91  & 1.08  & 25.8  & -   & 19.23 \\
YOLOX-Tiny  & 416   & 5.06  & 6.45  & 32.8  & -   & 32.77 \\
\hline
YOLOv5n  & 640   & 1.9  & 4.5  & 28.4  & 46.0   & 40.35 \\
YOLOv5s  & 640   & 7.2  & 16.5  & 37.2  & 56.0   & 78.05 \\
\hline
PP-PicoDet-ShuffleNetV2  & 416   & 1.17  & 1.53  & 30.0  & 44.6   & 15.06 $\vert$ \textbf{10.63$^\divideontimes$} \\
PP-PicoDet-MV3-large-1x  & 416   & 3.55  & 2.80   & 35.6  & 52.0   & 20.71 $\vert$ \textbf{17.88$^\divideontimes$} \\
PP-PicoDet-LCNet-1.5x  & 416   & 3.10 & 3.85   & 36.3  & 52.2   & 21.29 $\vert$ \textbf{20.8$^\divideontimes$} \\
\hline
\textbf{PP-PicoDet-S} & 320   & 0.99 & 0.73   & \textbf{27.1}    & \textbf{41.4}  & 8.13 $\vert$  \textbf{6.65$^\divideontimes$} \\
\textbf{PP-PicoDet-S} & 416   & 0.99 & 1.24  & \textbf{30.6}    & \textbf{45.5}  & 12.37  $\vert$ \textbf{9.82$^\divideontimes$} \\
\textbf{PP-PicoDet-M} & 416   & 2.15  & 2.50 & \textbf{34.3}    & \textbf{49.8}  & 17.39 $\vert$ \textbf{15.88$^\divideontimes$} \\
\textbf{PP-PicoDet-L} & 640   & 3.30  & 8.91   & \textbf{40.9}    & \textbf{57.6}  & 54.11 $\vert$ \textbf{50.55$^\divideontimes$} \\

\bottomrule[1pt]
\end{tabular}
\end{center}

\caption{Comparison of the speed and accuracy of different lightweight detectors on COCO \textit{val}. We use the NCNN\cite{ncnn2021} library for latency testing and add Paddle Lite\cite{paddlelite2021} inference latency(Marked as $^\divideontimes$) for the PP-PicoDet model. The latency of all models tested on Qualcomm Snapdragon$^\circledR$ 865(4*A77+4*A55) Processor with batch size of 1 and CPU threads of 4.}
\label{tab-state-of-the-art}
\end{table*}

\subsection{Label Assignment Strategy and Loss}
The label assignment of positive and negative samples has an essential impact on the object detectors. Most object detectors use fixed label assignment strategies. These strategies are straightforward. RetinaNet\cite{lin2017focal} directly divides positive and negative samples by the IoU of the anchor and ground truth. FCOS\cite{tian2019fcos} takes the anchors whose center point is inside the ground truth as positive samples, and 
YOLOv4\cite{bochkovskiy2020yolov4} and YOLOv5\cite{glenn_jocher_2021_5563715} select the location of the ground truth center point and its adjacent anchors as positive samples. ATSS\cite{atss} determines the positive and negative samples based on the statistical characteristics of the nearest anchors around the ground truth. The above-mentioned label assignment strategies are immutable in the global training process.
SimOTA is a label assignment strategy that changes continuously with the training process and achieves good results in YOLOX\cite{ge2021yolox}.  

We use SimOTA dynamic label assignment strategy to optimize our training process. SimOTA first determines the candidate area through the center prior, and then calculates the IoU of the predicted box and ground truth in the candidate area, and finally obtains parameter $\kappa$ by summing the n largest IoU for each ground truth. The cost matrix is obtained by directly calculating the loss for all predicted boxes and ground truth in the candidate area. For each ground truth, the anchors corresponding to the smallest $\kappa$ loss are selected and assigned as positive samples. The original SimOTA uses the weighted sum of CE loss and IoU loss to calculate the cost matrix. 
To align the cost in SimOTA and the objective function, we use the weighted sum of Varifocal loss and GIoU loss for cost matrix. The weight of GIoU loss is the $\lambda$, which is set to 6 as shown to best through our experiments. The specific formula is:
$$cost = loss_{vfl} +  \lambda \cdot loss_{giou} \eqno{(1)}$$
In the detector head, for classification, we use Varifocal loss to couple classification prediction and quality prediction. For regression, we use GIoU loss and Distribution Focal Loss. The formula is as follows:
$$loss = loss_{vfl} +2  \cdot loss_{giou}+0.25  \cdot loss_{dfl} \eqno{(2)}$$
In all the above formulas, $loss_{vfl}$ means Varifocal Loss, $loss_{giou}$ means GIoU loss, $loss_{dfl}$ means Distribution Focal Loss.

\subsection{Other Strategies}
In recent years, more activation functions have emerged that go beyond ReLU. Among these activation functions, H-Swish, a simplified version of the Swish activation function, is faster to compute and more mobile-friendly. We replace the activation function in detectors from ReLU to H-Swish. The performance improves significantly while keeping the inference time unchanged.

Different from linear step learning rate decay, cosine learning rate decay is exponentially decaying the learning rate. Cosine learning rate drops smoothly, which benefits the training process, especially when the batch size is large.

Too much data augmentation always increases the regularization effect and makes the training more difficult to converge for lightweight models. So in this work, we only use random flip, random crop and multi-scale resize for data augmentation in training.

\begin{table*}[!htbp]
\centering
\begin{center}
\begin{tabular}{c|c|c}
\toprule[1pt]
Model & Params(M) & mAP(0.5:0.95)\\
\midrule[1pt]
Base & 0.96  & 25.3 \\
 +CSP-PAN (3 feature maps)& 1.12 & 28.1\\
 +CSP-PAN (4 feature maps)& 1.17 & 29.1\\
 +Replace QFL with VFL& 1.17 & 29.2\\
 +Original SimOTA& 1.17 & 29.2\\
  +SimOTA with modified cost matrix& 1.17 & 30.0\\
   +Replace backbone with ESNet-0.75x& 0.99 &29.7\\
   +Replace LeakyRelu with H-Swish& \textbf{0.99} & \textbf{30.6}\\
\bottomrule[1pt]
\end{tabular}
\end{center}
\caption{Different configurations of ablation experiments in PP-PicoDet-S.}
\label{picodet_s_ablation}
\end{table*}

\section{Experiments}

\subsection{Implementation Details}

For training, we use stochastic gradient descent (SGD) with momentum of 0.9 and weight decay of 4e-5. The cosine decay learning rate scheduling strategy is adopted with initial learning rate  of 0.1. The batch size is 80x8 by default on 8x32G V100 GPU devices. We train 300 epochs, which costs 2 to 3 days.
All experiments are trained on COCO-2017\cite{mscoco} training set with 80 classes and 118k images, and are evaluated on COCO-2017 validation set with 5000 images using the standard COCO AP metric of a single scale. 
Exponential Moving Average (EMA) heavily utilizes recent information and
maintains long-term influence intuitively. Lightweight models are more likely to trap in local optima and are harder to converge. Therefore, we introduce a mechanism that works like regularization, named Cycle-EMA, to reset the content of the history, governed by a forget step. 


For the architecture search task, the settings of all hyper-parameters and datasets for supernet training are the same as the original model, detailed in the following section. We use L2 norm gradient clipping to avoid exploding gradients. Another difference is that we train eight candidates for each step since our search space is large.

\subsection{Ablation Study}
The results of all our ablation experiments are shown in Table \ref{picodet_s_ablation}. All experimental results are on COCO-2017 validation set.

\textbf{CSP-PAN.} 
We first get our base model similar to NanoDet, the backbone adopts ShuffleNetV2-1x, the neck adopts PAN without convolution, the loss adopts standard GFL\cite{li2020generalized} loss, and the label assignment strategy adopts ATSS\cite{atss}. All activation functions use LeakyRelu. The resulting mAP (0.5:0.95) is 25.3.
Further, we adopt the CSP-PAN structure. The feature map scale is 3. The mAP (0.5:0.95) is increased to 28.1.
Finally, we add a feature map scale to the top of CSP-PAN. Just like the final structure of our CSP-PAN, the number of parameters increases by less than 50K. The mAP (0.5:0.95) is further improved to 29.1. Results are shown in Table \ref{picodet_s_ablation}.

\textbf{Loss.} 
We compare the effects of Varifocal Loss (VFL) and Quality Focal Loss (QFL) under the same configuration in the previous section. The two are close, and the effect of Varifocal Loss is only slightly better than that of Quality Focal Loss. Replacing QFL with VFL, the mAP (0.5:0.95) is improved to 29.2 from 29.1. Results are shown in Table \ref{picodet_s_ablation}.

\textbf{Label Assignment Strategy.} 
Under the same configuration in the previous section, we replace ATSS with the original SimOTA and our modified SimOTA. We find that the larger the n, the worse the effect. The parameter n is then set to 10. The performance of ATSS is almost the same as the original SimOTA. The mAP (0.5:0.95) of our modified SimOTA achieves 30.0. Results are shown in Table \ref{picodet_s_ablation}.

We further compare the effect of SimOTA when Varifocal Loss and GIoU loss have different weights. We change the $\lambda$ of formula (1) to perform ablation experiments. The results are shown in Table \ref{simota_ablation}. When the weight of GIoU loss is 6, the best result is obtained.

\begin{table}[!htbp]
\centering
\begin{center}
\begin{tabular}{c|c}
\toprule[1pt]
$\lambda$& mAP(0.5:0.95) \\
\midrule[1pt]
5& 29.8\\
 6& 30.0\\
 7& 29.8\\
\bottomrule[1pt]
\end{tabular}
\end{center}
\caption{Different $\lambda$ on SimOTA with modified cost matrix.}
\label{simota_ablation}
\end{table}

\textbf{ESNet Backbone.}
We compare the performance of ESNet-1x and the original ShuffleNetV2-1.5x on ImageNet-1k. Table  \ref{esnet_ablation} shows that with less inference time, ESNet achieved higher accuracy.

\begin{table}[!htbp]
\centering
\begin{center}
\begin{tabular}{c|c|c|c}
\toprule[1pt]
model & \makecell[c]{FLOPs\\ (M)} & \makecell[c]{Latency\\(ms)} & \makecell[c]{Top-1 Acc\\(\%)} \\
\midrule[1pt]
ShuffleNetV2-1.5x & 301 & 7.56 & 71.6\\
\textbf{ESNet-1x} & \textbf{197} & \textbf{7.35} & \textbf{73.9} \\

\bottomrule[1pt]
\end{tabular}
\end{center}
\caption{Comparison of the ShuffleNetV2 and ESNet (use H-Swish)  with batch size of 1 and with CPU threads of 4.}
\label{esnet_ablation}
\end{table}

We also compare the performance of the original model and our searched model, and the results are shown in Table \ref{nas}. The searched model under latency constraint only decreases by 0.2\% mAP with mobile CPU inference time speed up by 41.5\% (54.9\% using Paddle Lite).
We replace the backbone of the detector mentioned above with ESNet-0.75x, the number of parameters is reduced by nearly 200K, mAP (0.5:0.95) is finally to 29.7. Results
are shown in Table \ref{picodet_s_ablation}.

\begin{table}[!htbp]
\centering
\begin{center}
\begin{tabular}{c|c|c|c}
\toprule[1pt]
model & \makecell[c]{Params\\(M)} & \makecell[c]{Latency\\(ms)} & \makecell[c]{mAP\\(0.5:0.95)}\\
\midrule[1pt]
original & 2.35 & 24.6 & 34.5\\
\textbf{searched} & \textbf{2.15 (-9.3\%)} & \textbf{17.39 (-41.5\%)} & 34.3 (-0.2)\\

\bottomrule[1pt]
\end{tabular}
\end{center}
\caption{Comparison of the designed model and NAS searched model with batch size of 1 and CPU threads of 4.}
\label{nas}
\end{table}

\textbf{H-Swish Activation Function.} 
Finally, we replace the LeakyRelu with H-Swish for all activation functions, mAP (0.5:0.95) is finally increased to 30.6. Results
are shown in Table \ref{picodet_s_ablation}.

\subsection{Comparison with the SOTA}
From Table \ref{tab-state-of-the-art}, we can see that our models far exceed all YOLO models in accuracy and speed. These achievements are mainly owing to the following improvements: (1) our neck is lighter than the neck of the YOLO series, so that backbone and head can be assigned more weights. (2) the combination of our Varifocal loss dealing with class imbalance, dynamic and learnable sample assignment, and regression method based on FCOS performs better in lightweight models. With the same amount of parameters, both the mAP and latency of PP-PicoDet-S surpass the YOLOX-Nano and NanoDet. Both the mAP and latency of PP-PicoDet-L exceed the YOLOv5s. Due to the more efficient convolution operator optimization by assembly language, we find that the inference testing performance of our models is even better when using Paddle Lite than using NCNN. In conclusion, our models are ahead of the SOTA models to a large extent. 

\section{Conclusion and Future Work}
We offer a new series of lightweight object detectors, which have superior performance on object detection for mobile devices. To our best knowledge, our PP-PicoDet-S model is the first model with mAP (0.5:0.95) surpassing 30, while keeping 1M parameters and 100+ FPS on ARM CPU. Moreover, the mAP (0.5:0.95) of our PP-PicoDet-L model surpasses 40 with only 3.3M parameters. In the future, we will continue to investigate new techniques to provide more detectors with high accuracy and efficiency.

\section{Acknowledgments}
This work is supported by the National Key Research and Development Project of China (2020AAA0103503).


{\small
\bibliographystyle{unsrt}
\bibliography{egbib}

\begin{thebibliography}{10}

\bibitem{redmon2018yolov3}
Joseph Redmon and Ali Farhadi.
\newblock Yolov3: An incremental improvement.
\newblock {\em arXiv preprint arXiv:1804.02767}, 2018.

\bibitem{bochkovskiy2020yolov4}
Alexey Bochkovskiy, Chien-Yao Wang, and Hong-Yuan~Mark Liao.
\newblock Yolov4: Optimal speed and accuracy of object detection.
\newblock {\em arXiv preprint arXiv:2004.10934}, 2020.

\bibitem{glenn_jocher_2021_5563715}
Glenn Jocher, Alex Stoken, Ayush Chaurasia, Jirka Borovec, NanoCode012, TaoXie,
  Yonghye Kwon, Kalen Michael, Liu Changyu, Jiacong Fang, Abhiram V, Laughing,
  tkianai, yxNONG, Piotr Skalski, Adam Hogan, Jebastin Nadar, imyhxy, Lorenzo
  Mammana, AlexWang1900, Cristi Fati, Diego Montes, Jan Hajek, Laurentiu
  Diaconu, Mai~Thanh Minh, Marc, albinxavi, fatih, oleg, and wanghaoyang0106.
\newblock {ultralytics/yolov5: v6.0 - YOLOv5n 'Nano' models, Roboflow
  integration, TensorFlow export, OpenCV DNN support}, October 2021.

\bibitem{ge2021yolox}
Zheng Ge, Songtao Liu, Feng Wang, Zeming Li, and Jian Sun.
\newblock Yolox: Exceeding yolo series in 2021.
\newblock {\em arXiv preprint arXiv:2107.08430}, 2021.

\bibitem{tian2019fcos}
Zhi Tian, Chunhua Shen, Hao Chen, and Tong He.
\newblock Fcos: Fully convolutional one-stage object detection.
\newblock In {\em Proceedings of the IEEE/CVF international conference on
  computer vision}, pages 9627--9636, 2019.

\bibitem{rangiLyu2021nanodet}
{NanoDet} Authors.
\newblock {NanoDet}.
\newblock \url{https://github.com/RangiLyu/nanodet}, 2021.

\bibitem{li2020generalized}
Xiang Li, Wenhai Wang, Lijun Wu, Shuo Chen, Xiaolin Hu, Jun Li, Jinhui Tang,
  and Jian Yang.
\newblock Generalized focal loss: Learning qualified and distributed bounding
  boxes for dense object detection.
\newblock {\em arXiv preprint arXiv:2006.04388}, 2020.

\bibitem{zhang2021varifocalnet}
Haoyang Zhang, Ying Wang, Feras Dayoub, and Niko Sunderhauf.
\newblock Varifocalnet: An iou-aware dense object detector.
\newblock In {\em Proceedings of the IEEE/CVF Conference on Computer Vision and
  Pattern Recognition}, pages 8514--8523, 2021.

\bibitem{zheng2020distance}
Zhaohui Zheng, Ping Wang, Wei Liu, Jinze Li, Rongguang Ye, and Dongwei Ren.
\newblock Distance-iou loss: Faster and better learning for bounding box
  regression.
\newblock In {\em Proceedings of the AAAI Conference on Artificial
  Intelligence}, volume~34, pages 12993--13000, 2020.

\bibitem{shufflenetv2}
Ningning Ma, Xiangyu Zhang, Hai-Tao Zheng, and Jian Sun.
\newblock Shufflenet v2: Practical guidelines for efficient cnn architecture
  design.
\newblock In {\em Proceedings of the European conference on computer vision
  (ECCV)}, pages 116--131, 2018.

\bibitem{ppdet2021}
PaddlePaddle Authors.
\newblock {PaddleDetection}, object detection and instance segmentation toolkit
  based on paddlepaddle.
\newblock \url{https://github.com/PaddlePaddle/PaddleDetection}, 2021.

\bibitem{girshick2015fast}
Ross Girshick.
\newblock Fast r-cnn.
\newblock In {\em Proceedings of the IEEE international conference on computer
  vision}, pages 1440--1448, 2015.

\bibitem{he2015spatial}
Kaiming He, Xiangyu Zhang, Shaoqing Ren, and Jian Sun.
\newblock Spatial pyramid pooling in deep convolutional networks for visual
  recognition.
\newblock {\em IEEE transactions on pattern analysis and machine intelligence},
  37(9):1904--1916, 2015.

\bibitem{fasterrcnn}
Shaoqing Ren, Kaiming He, Ross Girshick, and Jian Sun.
\newblock Faster r-cnn: Towards real-time object detection with region proposal
  networks.
\newblock In {\em Advances in neural information processing systems}, pages
  91--99, 2015.

\bibitem{fpn}
Tsung-Yi Lin, Piotr Doll{\'a}r, Ross Girshick, Kaiming He, Bharath Hariharan,
  and Serge Belongie.
\newblock Feature pyramid networks for object detection.
\newblock In {\em Proceedings of the IEEE conference on computer vision and
  pattern recognition}, pages 2117--2125, 2017.

\bibitem{maskrcnn}
Kaiming He, Georgia Gkioxari, Piotr Doll{\'a}r, and Ross Girshick.
\newblock Mask r-cnn.
\newblock In {\em Proceedings of the IEEE international conference on computer
  vision}, pages 2961--2969, 2017.

\bibitem{cai2018cascade}
Zhaowei Cai and Nuno Vasconcelos.
\newblock Cascade r-cnn: Delving into high quality object detection.
\newblock In {\em Proceedings of the IEEE conference on computer vision and
  pattern recognition}, pages 6154--6162, 2018.

\bibitem{liu2016ssd}
Wei Liu, Dragomir Anguelov, Dumitru Erhan, Christian Szegedy, Scott Reed,
  Cheng-Yang Fu, and Alexander~C Berg.
\newblock Ssd: Single shot multibox detector.
\newblock In {\em European conference on computer vision}, pages 21--37.
  Springer, 2016.

\bibitem{redmon2017yolo9000}
Joseph Redmon and Ali Farhadi.
\newblock Yolo9000: better, faster, stronger.
\newblock In {\em Proceedings of the IEEE conference on computer vision and
  pattern recognition}, pages 7263--7271, 2017.

\bibitem{long2020pp}
Xiang Long, Kaipeng Deng, Guanzhong Wang, Yang Zhang, Qingqing Dang, Yuan Gao,
  Hui Shen, Jianguo Ren, Shumin Han, Errui Ding, and Shilei Wen.
\newblock Pp-yolo: An effective and efficient implementation of object
  detector.
\newblock {\em arXiv preprint arXiv:2007.12099}, 2020.

\bibitem{huang2021pp}
Xin Huang, Xinxin Wang, Wenyu Lv, Xiaying Bai, Xiang Long, Kaipeng Deng,
  Qingqing Dang, Shumin Han, Qiwen Liu, Xiaoguang Hu, Dianhai Yu, Yanjun Ma,
  and Osamu Yoshie.
\newblock Pp-yolov2: A practical object detector, 2021.

\bibitem{redmon2016you}
Joseph Redmon, Santosh Divvala, Ross Girshick, and Ali Farhadi.
\newblock You only look once: Unified, real-time object detection.
\newblock In {\em Proceedings of the IEEE conference on computer vision and
  pattern recognition}, pages 779--788, 2016.

\bibitem{law2018cornernet}
Hei Law and Jia Deng.
\newblock Cornernet: Detecting objects as paired keypoints.
\newblock In {\em Proceedings of the European conference on computer vision
  (ECCV)}, pages 734--750, 2018.

\bibitem{duan2019centernet}
Kaiwen Duan, Song Bai, Lingxi Xie, Honggang Qi, Qingming Huang, and Qi~Tian.
\newblock Centernet: Keypoint triplets for object detection.
\newblock In {\em Proceedings of the IEEE/CVF International Conference on
  Computer Vision}, pages 6569--6578, 2019.

\bibitem{zhou2019objects}
Xingyi Zhou, Dequan Wang, and Philipp Kr{\"a}henb{\"u}hl.
\newblock Objects as points.
\newblock {\em arXiv preprint arXiv:1904.07850}, 2019.

\bibitem{atss}
Shifeng Zhang, Cheng Chi, Yongqiang Yao, Zhen Lei, and Stan~Z Li.
\newblock Bridging the gap between anchor-based and anchor-free detection via
  adaptive training sample selection.
\newblock In {\em Proceedings of the IEEE/CVF conference on computer vision and
  pattern recognition}, pages 9759--9768, 2020.

\bibitem{cai2020yolobile}
Yuxuan Cai.
\newblock {\em YOLObile: Real-time object detection on mobile devices via
  compression-compilation co-design}.
\newblock PhD thesis, Northeastern University, 2020.

\bibitem{mbv3}
Andrew Howard, Mark Sandler, Grace Chu, Liang-Chieh Chen, Bo~Chen, Mingxing
  Tan, Weijun Wang, Yukun Zhu, Ruoming Pang, Vijay Vasudevan, et~al.
\newblock Searching for mobilenetv3.
\newblock In {\em Proceedings of the IEEE International Conference on Computer
  Vision}, pages 1314--1324, 2019.

\bibitem{tan2019efficientnet}
Mingxing Tan and Quoc Le.
\newblock Efficientnet: Rethinking model scaling for convolutional neural
  networks.
\newblock In {\em International Conference on Machine Learning}, pages
  6105--6114. PMLR, 2019.

\bibitem{tan2019mnasnet}
Mingxing Tan, Bo~Chen, Ruoming Pang, Vijay Vasudevan, Mark Sandler, Andrew
  Howard, and Quoc~V Le.
\newblock Mnasnet: Platform-aware neural architecture search for mobile.
\newblock In {\em Proceedings of the IEEE/CVF Conference on Computer Vision and
  Pattern Recognition}, pages 2820--2828, 2019.

\bibitem{pham2018efficient}
Hieu Pham, Melody Guan, Barret Zoph, Quoc Le, and Jeff Dean.
\newblock Efficient neural architecture search via parameters sharing.
\newblock In {\em International Conference on Machine Learning}, pages
  4095--4104. PMLR, 2018.

\bibitem{brock2017smash}
Andrew Brock, Theodore Lim, James~M Ritchie, and Nick Weston.
\newblock Smash: one-shot model architecture search through hypernetworks.
\newblock {\em arXiv preprint arXiv:1708.05344}, 2017.

\bibitem{ghiasi2019fpn}
Golnaz Ghiasi, Tsung-Yi Lin, and Quoc~V Le.
\newblock Nas-fpn: Learning scalable feature pyramid architecture for object
  detection.
\newblock In {\em Proceedings of the IEEE/CVF Conference on Computer Vision and
  Pattern Recognition}, pages 7036--7045, 2019.

\bibitem{chen2019detnas}
Yukang Chen, Tong Yang, Xiangyu Zhang, Gaofeng Meng, Chunhong Pan, and Jian
  Sun.
\newblock Detnas: Neural architecture search on object detection.
\newblock {\em arXiv preprint arXiv:1903.10979}, 1(2):4--1, 2019.

\bibitem{xiong2021mobiledets}
Yunyang Xiong, Hanxiao Liu, Suyog Gupta, Berkin Akin, Gabriel Bender, Yongzhe
  Wang, Pieter-Jan Kindermans, Mingxing Tan, Vikas Singh, and Bo~Chen.
\newblock Mobiledets: Searching for object detection architectures for mobile
  accelerators.
\newblock In {\em Proceedings of the IEEE/CVF Conference on Computer Vision and
  Pattern Recognition}, pages 3825--3834, 2021.

\bibitem{cui2021pplcnet}
Cheng Cui, Tingquan Gao, Shengyu Wei, Yuning Du, Ruoyu Guo, Shuilong Dong, Bin
  Lu, Ying Zhou, Xueying Lv, Qiwen Liu, Xiaoguang Hu, Dianhai Yu, and Yanjun
  Ma.
\newblock Pp-lcnet: A lightweight cpu convolutional neural network, 2021.

\bibitem{senet}
Jie Hu, Li~Shen, and Gang Sun.
\newblock Squeeze-and-excitation networks.
\newblock In {\em Proceedings of the IEEE conference on computer vision and
  pattern recognition}, pages 7132--7141, 2018.

\bibitem{ghostnet}
Kai Han, Yunhe Wang, Qi~Tian, Jianyuan Guo, Chunjing Xu, and Chang Xu.
\newblock Ghostnet: More features from cheap operations.
\newblock In {\em Proceedings of the IEEE/CVF Conference on Computer Vision and
  Pattern Recognition}, pages 1580--1589, 2020.

\bibitem{imagenet}
Jia Deng, Wei Dong, Richard Socher, Li-Jia Li, Kai Li, and Li~Fei-Fei.
\newblock Imagenet: A large-scale hierarchical image database.
\newblock In {\em 2009 IEEE conference on computer vision and pattern
  recognition}, pages 248--255. Ieee, 2009.

\bibitem{mscoco}
Tsung-Yi Lin, Michael Maire, Serge Belongie, James Hays, Pietro Perona, Deva
  Ramanan, Piotr Doll{\'a}r, and C~Lawrence Zitnick.
\newblock Microsoft coco: Common objects in context.
\newblock In {\em European conference on computer vision}, pages 740--755.
  Springer, 2014.

\bibitem{liu2018path}
Shu Liu, Lu~Qi, Haifang Qin, Jianping Shi, and Jiaya Jia.
\newblock Path aggregation network for instance segmentation.
\newblock In {\em Proceedings of the IEEE conference on computer vision and
  pattern recognition}, pages 8759--8768, 2018.

\bibitem{ncnn2021}
NCNN Authors.
\newblock {NCNN}.
\newblock \url{https://github.com/Tencent/ncnn}, 2021.

\bibitem{paddlelite2021}
PaddlePaddle Authors.
\newblock {Paddle Lite}, multi-platform high performance deep learning
  inference engine.
\newblock \url{https://github.com/PaddlePaddle/Paddle-Lite}, 2021.

\bibitem{lin2017focal}
Tsung-Yi Lin, Priya Goyal, Ross Girshick, Kaiming He, and Piotr Doll{\'a}r.
\newblock Focal loss for dense object detection.
\newblock In {\em Proceedings of the IEEE international conference on computer
  vision}, pages 2980--2988, 2017.

\end{thebibliography}
}

\end{document}